\documentclass{ifacconf}

\usepackage{graphicx}      
\usepackage{natbib}

\usepackage{amsmath, amsfonts, amssymb}

\makeatletter
\let\old@BIBLABEL\@BIBLABEL
\def\@BIBLABEL#1{\old@BIBLABEL{#1}\kern\z@}
\makeatother
\usepackage[colorlinks=true,linkcolor=blue,citecolor=blue,urlcolor=cyan]{hyperref}

\usepackage{url}
\usepackage{fancyhdr}
\begin{document}
\begin{frontmatter}

\title{Motion Planning for Mobile Manipulators Navigating Doorways via Model Predictive Control} 

\thanks[footnoteinfo]{This work was sponsored by AlphaZ Inc. during the author's internship in collaboration with their research team.}

\author[First]{Kasra Sinaei} 
\author[Second]{Kasun Weerakoon Kulathun Mudiyanselage} 
\author[Second]{Christopher Bradley}
\author[Second]{Seyed Abolfazl Fakoorian}
\author[First]{Donald Ebeigbe}

\address[First]{The Pennsylvania State University, University Park, PA 16802 USA (e-mail: kasra@psu.edu, donald@psu.edu).}
\address[Second]{AlphaZ Inc., 
   Baldwin Park, CA 91706 USA (e-mail: kasun@alpha-z.ai, cbrad@csail.mit.edu, sed@alpha-z.ai).}
\begin{abstract}                
Navigating doorways is a fundamental capability for mobile manipulators operating in human environments, requiring coordinated motion between the mobile base and manipulator arm. This paper presents a motion planning framework that generates dynamically feasible and collision-free trajectories for autonomously opening and traversing both push and pull doors. The proposed method formulates the robot and door as a coupled dynamical system within a nonlinear Model Predictive Control (MPC) optimization framework. Manipulation feasibility is enforced through a penalty-based constraint, avoiding explicit arm kinematic modeling in the planner. Simulations and a hardware experiment demonstrate that the approach successfully plans feasible trajectories for door traversal.
\end{abstract}


\end{frontmatter}
\thispagestyle{fancy}
\pagestyle{empty} 
\section{Introduction}
Mobile manipulators are a versatile robotic platform suitable for tasks such as material handling, object transfer, and package delivery, and they have the potential to aid individuals with disabilities (\cite{sandakalum2022motion, thakar2020manipulator, nanavati2023physically}).

They are currently deployed in a wide variety of environments, including warehouses, factories, offices, and hotels (\cite{thakar2023survey, ghodsian2023mobile}). The key characteristic of the mobile manipulator design that gives it a huge advantage over the conventional static robotic arms is the drastic increase in its workspace and operating area, which brings more flexibility to the robot and allows it to execute more complex tasks compared to simple mobile robots and conventional static arms (\cite{leve2025ling}). One skill in particular that allows a mobile manipulator to explore and act across human-designed spaces, which we address here, is opening and traversing doors.

The special design of mobile manipulators introduces some challenges with motion planning and control of the robot. To extract the full potential of the system, one needs to efficiently plan the motion of the combined mobile base and manipulator arm, and also effectively track the reference end-effector and base positions. Note that this article addresses only the motion-planning problem.

Autonomous door opening and traversal with robots is an interesting problem that involves a coordinated series of actions executed with the robotic arm and a mobile base. It has been studied by researchers and engineers since the 1990s (\cite{nagatani1995experiment}), and has remained an intriguing robotics problem to date (\cite{xiong2024adaptive}). Model predictive control (MPC) is a well-established optimal trajectory planning method compatible with a wide range of systems, including linear time invariant (LTI) plants, nonlinear dynamic systems, hybrid and switched systems (\cite{borrelli2017predictive}). In the next section, we'll briefly highlight some of the bold research addressing the motion planning problem of mobile robots traversing doors and highlight recent advances made in this context. 

\subsection{Related Works}
A common approach towards traversing a closed door with a mobile manipulator involves detecting and estimating the parameters of the door, planning a feasible collision-free path for the robot, and then executing the planned path with the motion controllers and kinematic solvers. \cite{nagatani1995experiment} broke down the door-opening operations into several sub-tasks and proposed a planning method based on the door parameters. \cite{chitta2010planning} proposed a novel approach that utilizes conventional search-based planners for scenarios that involve door traversing and demonstrated their approach with a single-arm mobile manipulator. \cite{jang2023motion} further improved the method proposed by Chitta by introducing an integer variable into the problem called the area-indicator, which encodes the motion of the door into the dynamics of the mobile base, and used SBPL (search-based planning library) to solve for the optimal path. \cite{thamrongaphichartkul2024enhancing} discussed the technical details of employing behavior trees instead of traditional finite-state machines (FSM) for the autonomous door traversing task and showed some hardware and simulation results of their proposed framework based on behavior trees. Formulating the door opening problem as an optimization problem and solving it in real-time is also a popular approach (\cite{wu2024real}); \cite{reister2022combining} combined the manipulation task and navigation task of mobile manipulators to solve for time-efficient object placement via mobile manipulators. 

MPC is a popular tool for motion planning and trajectory generation. Previous studies have demonstrated its ability to solve motion planning problems (\cite{kiani2024learning, zheng2025robust}). Motion planning for door traversal is not an exception in this case; \cite{lee2020aerial} used MPC for motion control of an aerial mobile manipulator opening a hinged door; the authors argued that the MPC facilitates the formulation of collision avoidance of the robot-door pair and other state constraints and their integration into the control problem. \cite{stuede2019door} integrated a feedforward task controller, a grasp controller, and a navigation framework for an impedance-controlled mobile manipulator that can autonomously traverse doors. \cite{ma2018optimal} studied the energy consumption of mobile manipulators opening doors and formulated an energy cost function to solve for energy-efficient paths. Whole-body control and whole-body planning of the mobile manipulators are intensely studied and are tackled with search-based algorithms like \cite{thakar2020manipulator}, optimization-based methods like \cite{leve2025ling}, imitation learning (\cite{fu2024mobile}), and emerging vision-language-action models (\cite{wu2025momanipvla}).  Note that practical implementations of the previously mentioned approaches require some degree of robot state estimation and door or door handle detection (\cite{wang2025doorbot}). In this work, we mainly focus on the motion planning of the door traversing problem, while building upon the essential localization and detection frameworks from state-of-the-art open source solutions. 

Unlike search-based and behavior-tree approaches that organize door traversal as discrete stages, our planner jointly optimizes mobile-base motion and door opening in a single continuous nonlinear program. In contrast to full whole-body formulations, it retains a low-dimensional planning state by representing arm reachability with a soft workspace penalty and leaving arm motion to the IK controller; the same formulation handles both push and pull doors while enforcing geometric door-clearance constraints.

\subsection{Summary of Contributions}
The primary contributions of the proposed planner are summarized as follows:
\begin{itemize}
    \item {Coupled Robot-Door Dynamic Model:} We propose a unified nonlinear model that includes mobile-base motion and door opening in a shared planning state.
    \item {Continuous Collision-Aware Planner:} We formulate a nonlinear MPC with a reachability penalty and geometric door-clearance constraint, avoiding integer variables and full arm kinematics in the planner.
    \item {Implementation details:} We detail practical aspects of the proposed motion planning framework by showcasing some simulations and a hardware experiment.
\end{itemize}

The remainder of this paper is organized as follows. Section \ref{section_proposal} presents the system description and the mathematical formulation of the door traversal problem, detailing the proposed MPC optimization problem. Section \ref{section_results} validates the effectiveness of our approach through high-fidelity simulations and a hardware demonstration to justify practicality.
\section{Proposed Framework}
The main objective of this work is to plan a feasible path that allows the robot to approach a closed door, manipulate it to an open state while passing through the opening. The planner must be versatile enough to handle both push-style and pull-style doors. To make the problem tractable and without the loss of generality, we make some assumptions on the combined robot-door modeling.

\textbf{Modeling Assumptions:} 1) The door is modeled as a rigid body rotating about a fixed hinge, with its state defined by a single angle, $\theta_\text{door}$. 2) The environment consists of a corridor of known dimensions. Obstacles could be integrated into the planning algorithm via a series of optimization constraints. 3) The mobile base could follow an omnidirectional mobile robot dynamics or unicycle robot dynamics. Herein, we only demonstrate the use of unicycle robot dynamics, as we are using a differential drive mobile base in our simulations and hardware experiments. 4) For the sake of simplicity, we assume that the robot base is geometrically represented as a circle for collision checking. The door panel is represented as a line segment. 

Aforementioned assumptions are just facilitating the problem formulation and do not pose any limitations on the framework. 
\begin{figure}
    \centering
    \includegraphics[width=0.95\linewidth]{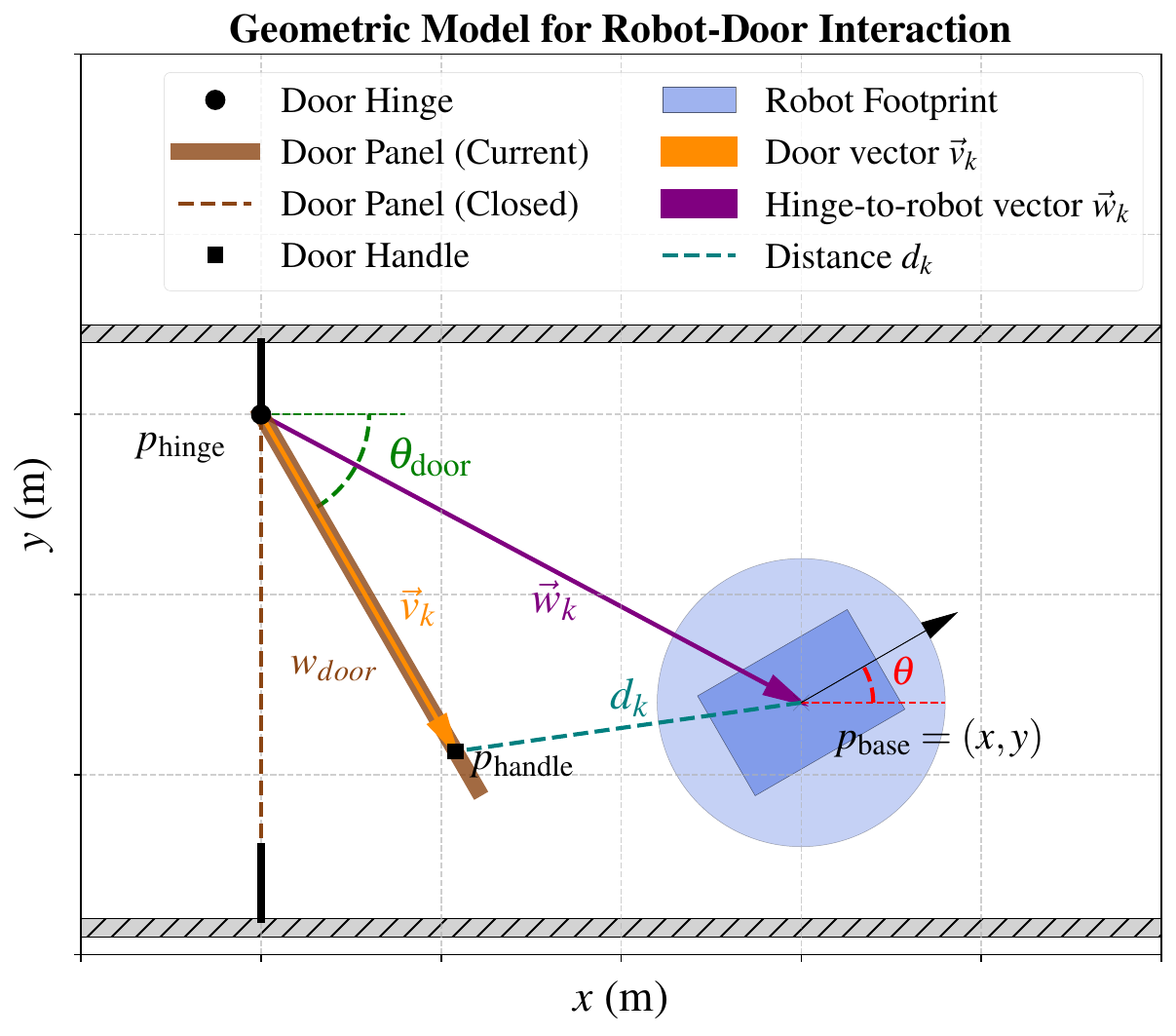}
    \caption{Geometric parameters of the mobile manipulator robot and its environment, including the door}
    \label{fig_parameters}
\end{figure}
\label{section_proposal}
\subsection{System Modeling}
Let us denote the configuration space of the robot and door in a 2D plane by the state vector $\mathbf{x} = \begin{bmatrix} x & y & \theta_\text{base} & \theta_\text{door}\end{bmatrix}^T$, where $x$ and $y$ are cartesian position of the mobile base in global frame, $\theta_\text{base}$ is the heading angle of the robot and the door opening angle is denoted with $\theta_\text{door}$. System inputs are denoted by $u = \begin{bmatrix} v & \omega_\text{base} & \omega_\text{door} \end{bmatrix}^T$. Consider the case that the mobile base follows the unicycle robot dynamics and the door is modeled as a single-integrator linear system ($\dot{\theta}_\text{door} = \omega_\text{door}$). To embed the manipulation behavior of the system into the model, we need to constrain the motion of the door such that its angular velocity is non-zero only if the mobile base is positioned close to the door handle. The resulting switched dynamical system is given by:
\begin{align} \label{eq_switched_dyn}
    \dot{\mathbf{x}} = \begin{bmatrix}
        \dot{x} \\ \dot{y} \\ \dot{\theta}_\text{base} \\ \dot{\theta}_\text{door}
    \end{bmatrix} = 
    \begin{bmatrix}
        v \cos(\theta_\text{base}) \\
        v \sin(\theta_\text{base}) \\
        \omega_\text{base} \\
        \delta_\text{door} \omega_\text{door}
        \end{bmatrix}
\end{align}
where
\begin{align}
        \delta_\text{door} =
        \begin{cases}
           1 & \text{if handle reachable} \\
           0 & \text{otherwise}
        \end{cases}
\end{align}
Intuitively, we can see that when the integer parameter $\delta_\text{door}$ is one, the single integrator dynamics of the door opening angle become effective; otherwise, the angular velocity remains zero. Although modern nonlinear optimization solvers are capable of solving problems with integer variables or those involving piece-wise-affine dynamics like our switched system, we prefer to remove the integer variable $\delta_\text{door}$ from the system dynamics and enforce this manipulation constraint through our cost function (soft constraint). Before eliminating it from the system dynamics, we demonstrate how the integer variable is defined and computed to make the formulation more comprehensive.
\begin{figure}[t]
    \centering
    \includegraphics[width=0.85\linewidth]{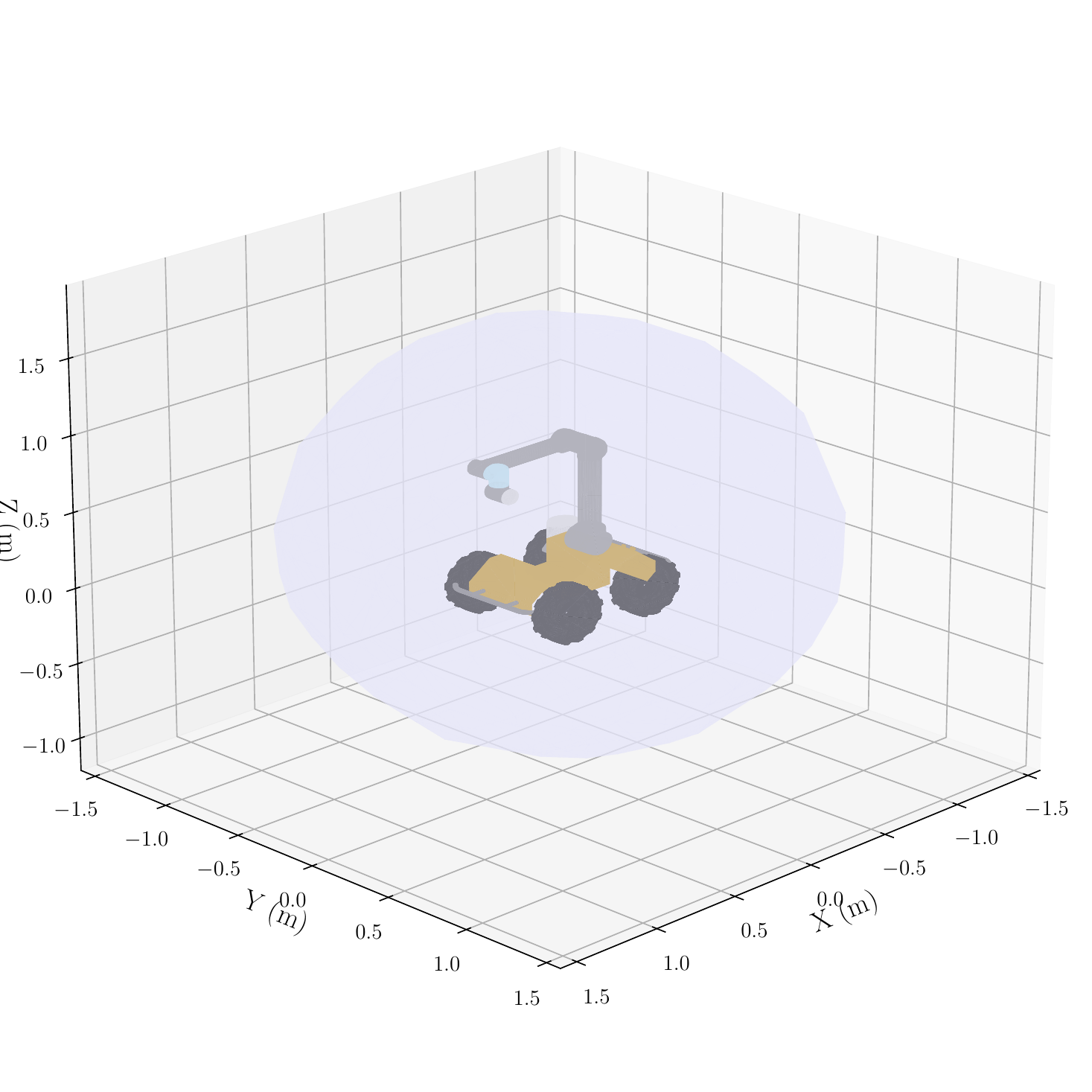}
    \caption{Workspace of a mobile manipulator robotic system consisting of a Husky A100 mobile base and an industrial UR10 robotics arm with 6 degrees of freedom. The dimensions of the convex hull of the arm workspace are important for determining the $r_{\min}$ and $r_{\max}$ in practice.}
    \label{fig_husky_ws}
\end{figure}
\subsection{Planner Formulation}
The value of variable $\delta_{door}$ is determined by the distance between the door handle and the mobile base of the robot. Depending on the size of the arm, there is a bounded area between two concentric spheres centered at the root link of the arm with radii $r_{\min}$ and $r_{\max}$, where the end-effector of the arm can reach any point in this region. If we denote the position of the mobile base in the k-th step by $p_{\text{base},k}=\begin{bmatrix}x_k & y_k \end{bmatrix}^T \in \mathbb{R}^2$, and the door handle position is denoted by $p_{\text{handle},k} \in \mathbb{R}^2$. The handle position $p_\text{handle}$ could be computed from the door opening angle and door hinge position $p_\text{hinge} \in \mathbb{R}^2$:
\begin{align}
    p_{\text{handle},k} =& p_\text{hinge} + \begin{bmatrix} d_\text{handle} \cos(\theta_{\text{door},k}) \\ d_\text{handle} \sin(\theta_{\text{door},k})  \end{bmatrix} \\
    d_k =& \|p_{\text{base},k} - p_{\text{handle},k}\|
\end{align}
The operator $\|.\|$ denotes the Euclidean norm of vectors and $d_\text{handle}$ refers to the distance from the door hinge to the door handle. The distance $d_k$ is the key distance to check if the door is manipulatable based on the current position of the mobile base and door angle. The integer variable is determined by:
\begin{align}\label{eq_integer_var}
    \begin{cases}
        \delta_\text{door} = 1 & \text{if}\; r_{\min} \le d_k \le r_{\max} \\  \delta_\text{door} = 0 & \text{otherwise} 
    \end{cases}
\end{align}
To simplify the problem, we define the manipulation penalty cost $C_m(\mathbf{x}_k)$ based on this distance to check if the door handle lies within the $r_{\min}-r_{\max}$ radius of the manipulator as shown in (\ref{eq_manipulation_cost}). We penalize this term with a high gain and the door angular velocity to guide the solver through feasible solutions without explicitly using $\delta_{door}$:
\begin{equation}\label{eq_manipulation_cost}
  \begin{aligned}
    C_\text{manip}(\mathbf{x}_k) &=
    \bigl(\max\!\{0,\, d_k^2 - r_{\max}^2\} \\
    &\qquad + \max\!\{0,\, r_{\min}^2 - d_k^2\}\bigr)
    |\omega_\text{door}|
  \end{aligned}
\end{equation}
We also need to define a constraint for avoiding collisions with the door panel. This constraint is enforced by ensuring the minimum distance between the robot's center and the door's line segment is greater than the base collision radius ($R_\text{base}$). One can formulate this by computing the position of the closest point to the robot's base on the line segment. Let us denote this point at time step $k$ by $p_{\text{closest},k}$. Define vector $\vec{v}_k = \mathbf{p}_{\text{handle},k} - \mathbf{p}_{\text{hinge}}$ and $\vec{w}_k = \mathbf{p}_{\text{base},k} - \mathbf{p}_{\text{hinge}}$ as illustrated in Figure \ref{fig_parameters}. Next, we find the scalar projection of $v_k$ onto $w_k$ by calculating a parameter $t_k$ via (\ref{eq_projection}). This parameter represents where the orthogonal projection of the robot's base falls along the infinite line containing the door vector:
\begin{align}\label{eq_projection}
    t_k = \frac{\vec{w}_k \cdot \vec{v}_k}{\vec{v}_k \cdot \vec{v}_k} = \frac{\vec{w}_k \cdot \vec{v}_k}{\|\vec{v}_k\|^2}
\end{align}
Now we can find the closest point of the door to the mobile base of the robot according to the value of the variable $t_k$:
\begin{align}
    {p}_{\text{closest},k} =
    \begin{cases}
    {p}_{\text{hinge}} & \text{if } t_k < 0 \\
    {p}_{\text{handle},k} & \text{if } t_k > 1 \\
    {p}_{\text{hinge}} + t_k \vec{v}_k & \text{if } 0 \le t_k \le 1
    \end{cases}
\end{align}
We can add a nonlinear constraint to the optimization problem to ensure that the minimum distance between the robot base and the closest point on the door panel is greater than the robot's radius; this approach can also be used for other convex or nonconvex obstacles surrounding the robot. With the manipulation cost defined in (\ref{eq_manipulation_cost}), one can formulate an MPC problem with quadratic state and control costs to solve for an optimal solution. Equation (\ref{eq_planner_cost}) shows the planner's cost function which consists of three terms, $C_\text{effort}(\mathbf{u}_k)$ ensures minimal control effort and optimality of the solution, $C_{m}(\mathbf{x}_k)$ which is multiplied by $w_m \in \mathbb{R}_+$ to enforce manipulation constraint, and finally the state cost $C_\text{final}(\mathbf{x}_N, \mathbf{x}_\text{goal})$ ensures that the system state will reach a desired goal state $\mathbf{x}_\text{goal}$ at the end of optimization horizon: 
\begin{align}
    &J = \sum_{k=0}^{N-1} \left( C_\text{effort}(\mathbf{u}_k) + w_m C_{m}(\mathbf{x}_k) \right) +  C_\text{final}(\mathbf{x}_N, \mathbf{x}_\text{goal}) \label{eq_planner_cost}\\
    &C_\text{final}(\mathbf{x}_N, \mathbf{x}_\text{goal}) = (\mathbf{x}_N - \mathbf{x}_\text{goal})^T \mathbf{Q}_f (\mathbf{x}_N - \mathbf{x}_\text{goal}) \\
    &C_\text{effort}(\mathbf{u}_k) = \mathbf{u}_k^T \mathbf{R} \mathbf{u}_k \\
    &C_{m}(\mathbf{x}_k) = w_\text{manip} \cdot C_\text{manip}(\mathbf{x}_k, \mathbf{u}_k)
\end{align}
\begin{figure}[t!]
    \centering
    \includegraphics[width=0.9\linewidth]{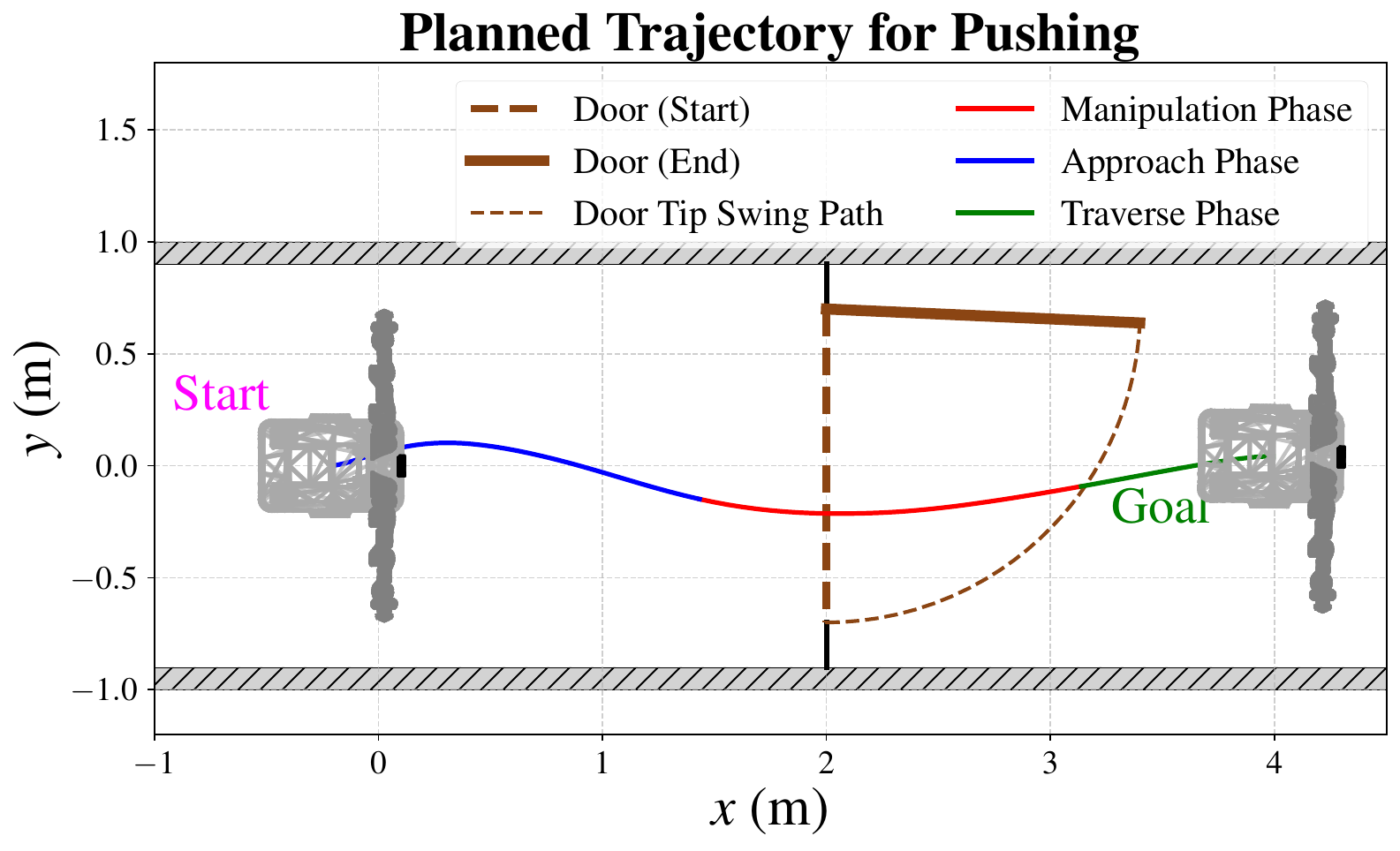}
    \caption{Planned trajectory for a push-type door.}
    \label{fig_push_path}
\end{figure}
We use subscripts $k$ and $N$ for indexing the state and control. The constraints of the optimization problem are essential for the collision avoidance with the door, and also for enforcing the maximum velocity of the robot. The final nonlinear MPC problem is formulated as:
\begin{align}
    \mathbf{X, U} = \arg&\min_{\mathbf{x}_k, \mathbf{u}_k} \quad J(X, U, \mathbf{x}_{goal}) \label{eq_planner_mpc} \\
    \text{s.t.} \quad & \mathbf{x}_{k+1} = \mathbf{x}_k + f(\mathbf{x}_k, \mathbf{u}_k) dt \\
    & \| \mathbf{p}_{\text{base},k} - \mathbf{p}_{\text{closest},k} \|^2 \ge (R_\text{base} + d_{s})^2 \label{eq_door_base_constrain}\\
    & \mathbf{u}_{\min} \le \mathbf{u}_k \le \mathbf{u}_{\max} \\
    & \mathbf{x_{\min} \le \mathbf{x}_k \le \mathbf{x}_{\max}} \label{eq_planner_state_bound}
\end{align}
where $f(x_k, u_k)$ is the system dynamics of the combined door-robot system described by (\ref{eq_switched_dyn}), where the dynamics of the fourth state follow a simple single-integrator instead of the switching condition. ($\dot \theta_\text{door} = \omega_\text{door}$), $d_s$ is a safe margin added to the mobile base radius to have a larger threshold for the collision check. The solution of the nonlinear program $(\mathbf{X}, \mathbf{U})$ includes a series of length $N$, including both the states $\mathbf{x}_k$ and controls $\mathbf{u}_k$.

\begin{figure}[t!]
    \centering
    \includegraphics[width=0.9\linewidth]{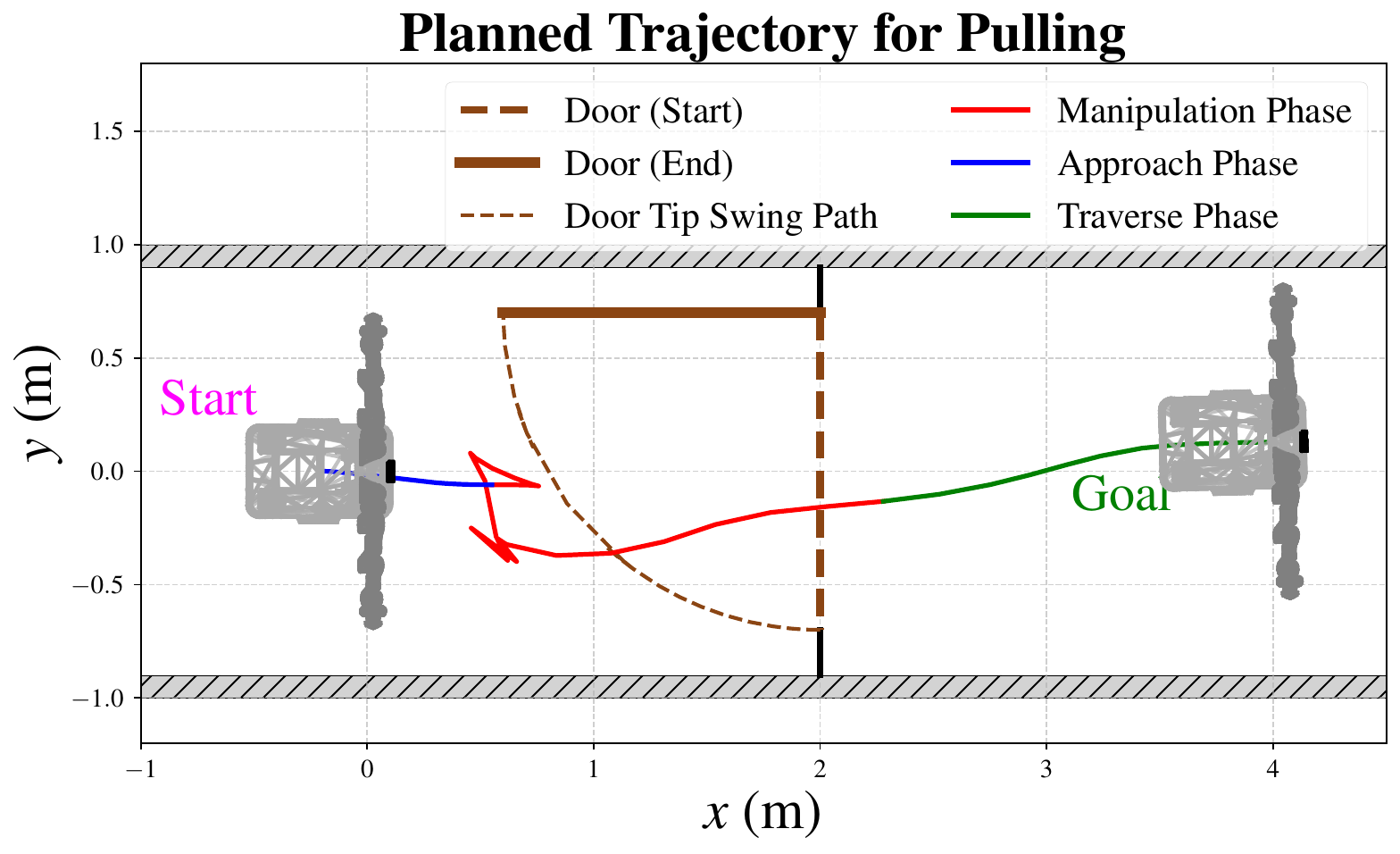}
    \caption{Planned path for a pull-type door.}
    \label{fig_pull_path}
\end{figure}

\begin{figure*}[t!]
    \centering
    \includegraphics[width=0.95\linewidth]{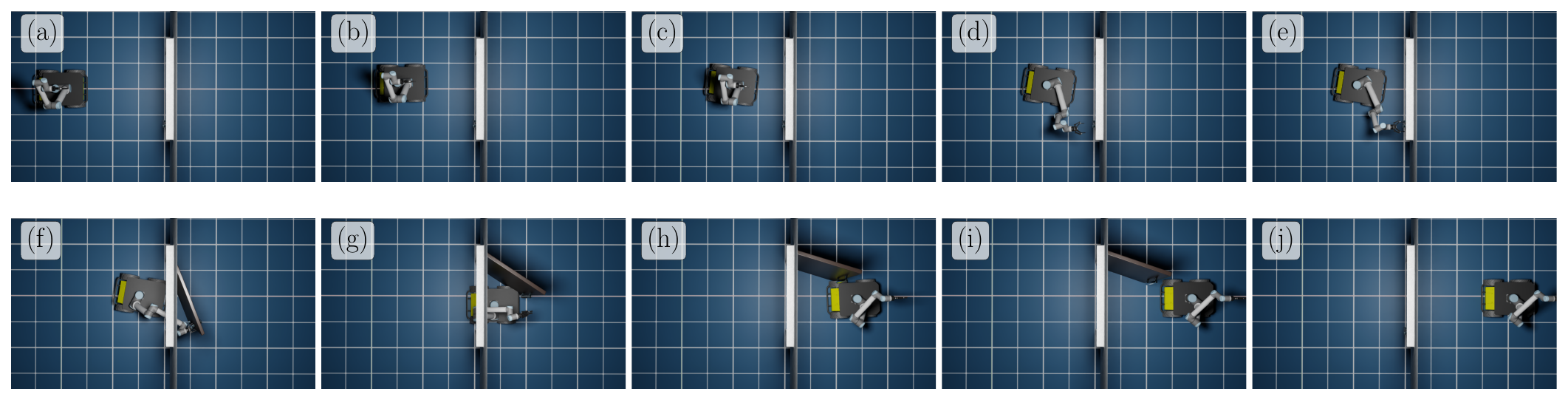}
    \caption{Simulation of a mobile manipulator in NVIDIA Isaac Sim; demonstration of MPC on push-type door.}
    \label{fig_sim_push}
\end{figure*}
\begin{figure*}[t!]
    \centering
    \includegraphics[width=0.95\linewidth]{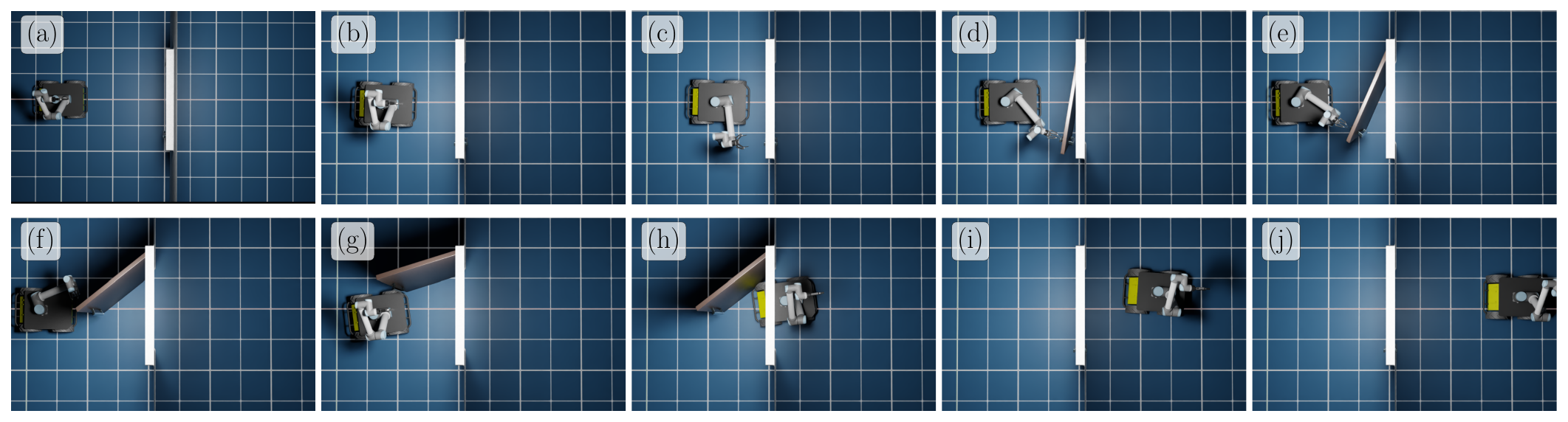}
    \caption{Simulation of a mobile manipulator in NVIDIA Isaac Sim; demonstration of MPC on pull-type door.}
    \label{fig_sim_pull}
\end{figure*}

\section{Simulation and Experiment}\label{section_results}
Trajectories generated by the nonlinear program are ensured to be feasible and collision-free. In this section, we briefly discuss the mobile base velocity controller and the kinematic solver of the manipulator arm. A whole-body controller for the robotic system is favorable; however, we used this hierarchical framework just to demonstrate simulation and experiments that justify the planner's practicality.

\textbf{Differential Inverse Kinematics:} We use the inverse kinematics solver proposed by \cite{pink}, which finds the differential IK weighted by task solution in real-time. This framework enables enforcing collision constraints via control barrier functions (CBFs). If we denote the 6-DoF arm configuration by $q \in \mathbb{R}^6$ and task errors by $e_\text{task}(q) = p_\text{task}^\text{des} - p_\text{task}(q)$, where $p_\text{task}^\text{des}$ is the desired task goal and $p_\text{task}(q)$ is the current task value for configuration $q$; then (\ref{eq_pink}) shows the quadratic program formulation of the IK solver. Subscript of \textit{task} is dropped for brevity, and $J_e(q)=\frac{\partial e}{\partial q}$ is the task Jacobian:
\begin{align}\label{eq_pink}
    v =& \arg\min \sum_\text{tasks} \|J_e(q)v + \alpha e(q) \|_{W_e}^2 \\
    \text{s.t.}& \qquad v_{\min} \le v \le v_{\max} \nonumber \\
    & \qquad\nabla h_i(q) v + \alpha_i(h_i(q)) \ge 0 \nonumber
\end{align}
The parameter $\alpha$ is a positive scalar gain that controls the convergence rate of the solver, and $W_e$ is the task weight matrix. The operator $\|.\|_W$ is the weighted sum of the norms with weight matrix $W$. The solution of the optimization problem is the velocity that we need to apply to each joint to get closer to task targets. To solve this problem for a mobile manipulator, we define one task for the mobile base and one task for the end-effector. The desired target for the base task is set according to the estimated position of the base coming from localization, and the target for the end-effector task is set from the desired position of the door handle computed from the planned door opening angle obtained by (\ref{eq_planner_mpc}). In the second constraint, $h_i$ denotes the barrier function, which serves as a safety filter for the inverse kinematic solver and is used to avoid collisions (see \cite{ames2016control} for more details on CBF), while $\alpha_i(.)$ is a class-$\mathcal{K}$ function (refer to \cite{khalil2002nonlinear} for definition). Self-collision barrier and wall barriers are used in the simulations and experiment.

\textbf{Base Controller:} The mobile base is controlled using a nonlinear MPC tracker based on unicycle dynamics. This MPC formulation will seamlessly generate a proper velocity $u_{b} = \begin{bmatrix} v & \omega_\text{base} \end{bmatrix}$ command for the mobile base. Define the base tracking error as $\mathbf{e}_{b,k} = \mathbf{x}_k - \mathbf{x}_{g,k}$ where $\mathbf{x}_k=\begin{bmatrix}x & y & \theta_\text{base} \end{bmatrix}^T$ and the $\mathbf{x}_g$ is the target position of the base which will be set according to the output of planner optimization problem (\ref{eq_planner_mpc}).
\begin{align}\label{eq_base_mpc}
    \mathbf{u}_\text{base} = & \arg \min  \sum_{k=1}^N \left(\mathbf{u}_{b,k}^TR_{b}\mathbf{u}_{b,k} + \mathbf{e}_{b,k}^TQ_{b}\mathbf{e}_{b,k}\right) \\
    &\text{s.t.}\:\: \mathbf{x}_{k+1} = f_\text{uni}\left(\mathbf{x}_k, \mathbf{u}_{b,k}\right) \forall k \in [1, ..., N]\nonumber \\
    & \mathbf{u}_{\min} \le \mathbf{u}_{\text{base},k} \le \mathbf{u}_{\max} \nonumber
\end{align}
where $f_\text{uni}(x_k, u_k)$ is the discretized dynamics of the unicycle robot (\cite{tzafestas2014mobile}) and $u_{\min}, u_{\max}$ are velocity limits of the mobile robot. The horizon of the MPC is also denoted by $N$.
\begin{figure*}
    \centering
    \includegraphics[width=0.95\linewidth]{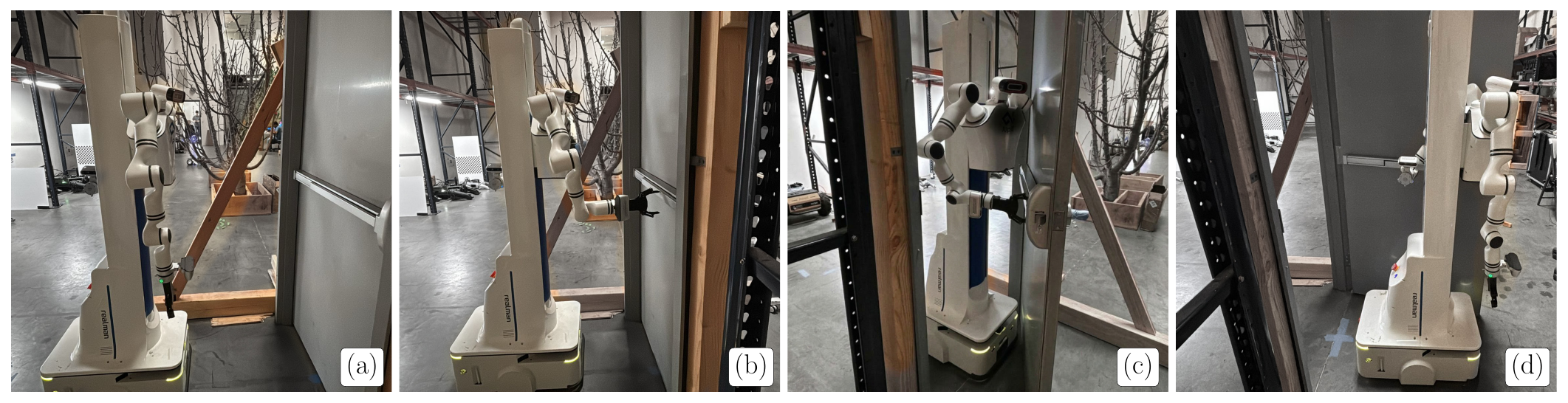}
    \caption{Snapshots of the hardware experiment using a commercial mobile manipulator. The robot is traversing a warehouse door using the path generated by the MPC. Video available at: \url{https://kassra-sinaei.github.io/mpc_doorbot/}}
    \label{fig_rm_push}
\end{figure*}

Quadratic program \eqref{eq_planner_mpc} is solved with the IPOPT solver through the CasADi library (\cite{andersson2019casadi}). The MPC problem only needs to be solved once and takes a few seconds to generate the entire door opening trajectory. The tracking controller of base velocity and the arm's kinematic solver work simultaneously to traverse waypoints generated by the planner. We used Pink-IK (\cite{pink}) to solve the differential IK problem \eqref{eq_pink} and IPOPT to solve \eqref{eq_base_mpc} in real-time. A UR-10 arm mounted on a Husky A100 robot is simulated in NVIDIA Isaac-Sim, and a Realman commercial mobile manipulator is used for the experiment. 

Figures \ref{fig_sim_push}-\ref{fig_rm_push} show snapshots of these simulations and experiments, in which the robot could successfully traverse the door and reach its designated goal position. More details and videos are available online at \url{https://kassra-sinaei.github.io/mpc_doorbot/}. The successful execution of these experiments highlights the framework's versatility in handling both push and pull door mechanisms seamlessly. By effectively integrating the MPC planner with real-time inverse kinematics and model predictive control, the system demonstrates a robust capability to translate complex, coupled dynamical plans into smooth, reliable hardware actions.

\section{Conclusion}
This paper presented a motion planning framework for mobile manipulators to autonomously open and traverse doors using a nonlinear MPC formulation. The robot and door were modeled as a coupled dynamical system, while manipulation feasibility and collision avoidance were incorporated through optimization costs and constraints. Simulation and hardware experiments demonstrated that the proposed planner can generate feasible trajectories for both push and pull door scenarios.
\bibliography{ifacconf}            

\end{document}